\documentclass[letterpaper, 10 pt, journal, twoside]{ieeetran}

\usepackage{etoolbox}
\makeatletter
\patchcmd{\@makecaption}
  {\scshape}
  {}
  {}
  {}
\makeatother

\IEEEoverridecommandlockouts                               

\usepackage{bm}

\usepackage{graphicx}                       
\usepackage{graphics}                       
\usepackage{epsfig}                         
\usepackage[tight,footnotesize]{subfigure}  
\graphicspath{./pics/RAL_fig_for_revise}
\graphicspath{./pics/RAL_new_fig}
\usepackage{amssymb,amsmath}
\usepackage{mdwmath}
\usepackage{commath}   
\usepackage{eqparbox}
\usepackage{mathtools}
\usepackage[utf8]{inputenc} 
\usepackage[english]{babel}
\usepackage{amsmath,amsfonts,amssymb}

\newcommand{\beq}{\begin{equation}}
\newcommand{\eeq}{\end{equation}}
\newcommand{\bear}{\begin{eqnarray}}
\newcommand{\bears}{\begin{eqnarray*}}
\newcommand{\eear}{\end{eqnarray}}
\newcommand{\eears}{\end{eqnarray*}}
\newcommand{\bdm}{\begin{displaymath}}
\newcommand{\edm}{\end{displaymath}}
\newcommand{\lba}{\left[\begin{array}}
\newcommand{\ear}{\end{array}\right]}

\usepackage{stfloats}                       
\usepackage{url} 
\usepackage{hyperref}
\usepackage{cite}                           
\usepackage[T1]{fontenc} 
\usepackage{tabularx}
\usepackage{diagbox} 
\usepackage{color}
\usepackage{multirow}
\usepackage[table]{xcolor}
\usepackage{longtable}
\usepackage{booktabs} 			
\usepackage{xcolor,colortbl}    

\usepackage{multicol}
\usepackage{graphicx}
\usepackage{placeins}

\usepackage{amssymb}
\usepackage{pifont}
\newcommand{\xmark}{\ding{55}}%
\title{Tactile Gym 2.0: Sim-to-real Deep Reinforcement Learning for Comparing Low-cost High-Resolution Robot Touch} 

\author{
Yijiong Lin$^{1}$, 
John Lloyd$^{1}$, 
Alex Church$^{1}$, 
Nathan F. Lepora$^{1}$ \\
Project Webpage: \url{https://sites.google.com/my.bristol.ac.uk/tactilegym2}
\thanks{
YL was supported by the China Scholarship Council (CSC)/University of Bristol joint-funded scholarship. AC was supported by an EPSRC CASE award sponsored by Google DeepMind. NL and JL were supported by a Leadership Award from the Leverhulme Trust on ‘A biomimetic forebrain for robot touch’ (RL-2016-39).
}
\thanks{$^{1}$ All authors are with the Department of Engineering Mathematics and Bristol Robotics Laboratory, University of Bristol, Bristol BS8 1UB, U.K. (email: \{yijiong.lin, j.lloyd, a.church, n.lepora\}@bristol.ac.uk)}
\thanks{Code is open-sourced at: \url{https://github.com/ac-93/tactile_gym}}

}

\begin{document}
\maketitle
\begin{abstract}

High-resolution optical tactile sensors are increasingly used in robotic learning environments due to their ability to capture large amounts of data directly relating to agent-environment interaction. However, there is a high barrier of entry to research in this area due to the high cost of tactile robot platforms, specialised simulation software, and sim-to-real methods that lack generality across different sensors. In this letter we extend the Tactile Gym simulator to include three new optical tactile sensors (TacTip, DIGIT and DigiTac) of the two most popular types, Gelsight-style (image-shading based) and TacTip-style (marker based). We demonstrate that a single sim-to-real approach can be used with these three different sensors to achieve strong real-world performance despite the significant differences between real tactile images. Additionally, we lower the barrier of entry to the proposed tasks by adapting them to an inexpensive 4-DoF robot arm, further enabling the dissemination of this benchmark. We validate the extended environment on three physically-interactive tasks requiring a sense of touch: object pushing, edge following and surface following. The results of our experimental validation highlight some differences between these sensors, which may help future researchers select and customize the physical characteristics of tactile sensors for different manipulations scenarios.
\end{abstract}

\section{INTRODUCTION}\label{sec:Intro}


%


\begin{figure}[t!]
  \centering
     \includegraphics[width=0.95\linewidth]{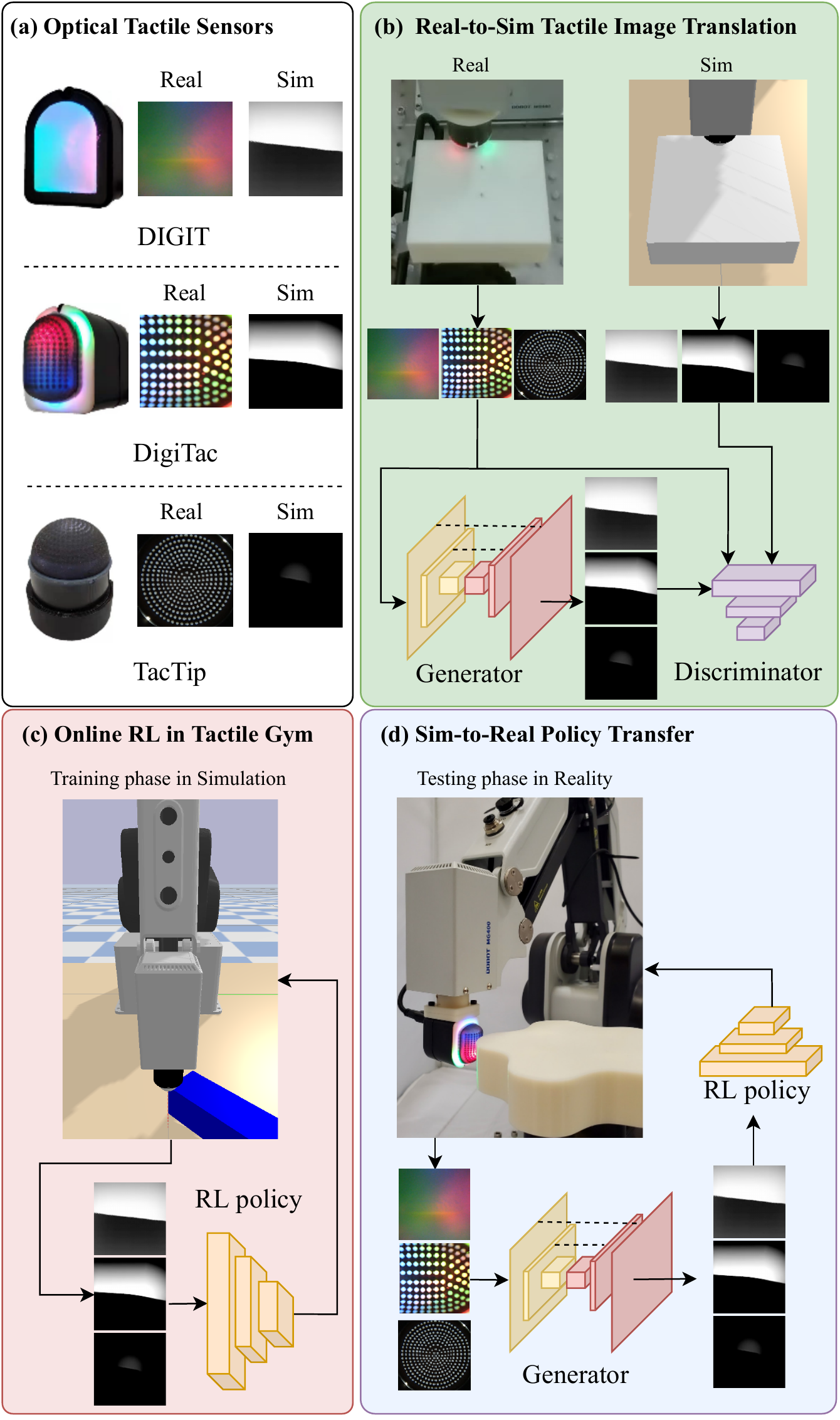}
    \caption{Overview of the tactile sim-to-real deep RL robotic system: (a)~Three low-cost high-resolution optical tactile sensors and the raw sensor images. (b) Training real-to-sim tactile image translation. (c) Policy learning in the Tactile Gym \cite{church_tactile_2021} with simulated tactile images from multiple integrated sensors. (d) A desktop robot equipped with the DigiTac performing the surface-following task by translating real tactile images to simulated images for the RL policy.}
  \label{fig:overview}
  \vspace{-1em}
\end{figure}

A plausible route to human-like robot dexterity is to combine deep learning with high-resolution tactile sensing, given the unprecedented recent advances in controlling robots with deep learning applied to robot vision~\cite{sunderhauf2018limits}. Moreover, the use of deep reinforcement learning (RL) would seem to offer the potential for learning complex manipulation tasks based on a reward information, which is both a mechanism for how humans acquire new skills and has achieved impressive results in simulated environments such as computer games. However, there are major challenges preventing tactile deep RL from being realised: (1) the lack of available and accessible tactile sensing technologies limits the research capacity available to develop RL methods for touch; (2) those labs that have expertise in fabricating tactile sensors tend to stay with the technology where they have expertise; (3) hence, approaches to tactile deep RL, e.g.~\cite{church_tactile_2021,dong2021tactile,xu2021towards}, stay confined within those labs, which is inefficient for progressing the field and opposite to the open culture that has benefitted AI research.

Meta AI researchers have developed and open-sourced a tactile robot learning platform called TACTO~\cite{wang2022tacto} and tactile processing libraries PyTouch~\cite{lambeta2021pytouch} for GelSight-based tactile sensors such as the low-cost, open-sourced DIGIT tactile sensor \cite{lambeta2020digit}. Meanwhile, researchers in Bristol Robotics Laboratory have developed a family of high-resolution biomimetic optical tactile sensors called the TacTip \cite{ward2018tactip,lepora2021soft}, alongside an open-sourced suite of learning environments called Tactile Gym that features highly-efficient tactile image rendering \cite{church_tactile_2021}. In consequence, sim-to-real policy transfer via tactile image translation enabled zero-shot performance on multiple exploration and manipulation tasks requiring tactile feedback, such as delicately tracing the edges and surfaces of complex objects and pushing objects to goal locations. 

The goal of this present research is to bring together this progress with GelSight-type sensors with the sim-to-real deep RL methods developed with the TacTip. Whilst it was previously claimed that the approach used in \cite{church_tactile_2021} should be applicable to ``a broad range of other high-resolution optical tactile sensors including sensors of the Gelsight type'', the approach was only demonstrated with the TacTip. Here we extend the approach to the DIGIT sensor (of Gelsight type), a reduced form-factor TacTip sensor and a DIGIT-TacTip hybrid sensor referred to as the DigiTac. We demonstrate strong real-world performance across these sensors despite the significantly different tactile images produced by each.

The main contributions of this work are as follows:\\
\noindent1) We extend the Tactile Gym \cite{church_tactile_2021} from one to three low-cost, high-resolution optical tactile sensors. To the best of our knowledge, this is the first work that integrates two widely-used yet fundamentally different styles of the optical tactile sensors: Gelsight-style (DIGIT) and TacTip-style (DigiTac and TacTip), into one platform. Such an integration validates and extends earlier results for this platform while making it more accessible and applicable to the wider research community.

\noindent2) With the extended Tactile Gym environments, we successfully learn deep RL policies for three physically-interactive tasks (edge following, surface following and object pushing), and transfer them into the real world without further tuning. By comparing task performance, we identify strengths and weaknesses of the tactile sensors in these different scenarios. As far as we know, this is the first empirical comparison of optical tactile sensors in a sim-to-real context, which we intend as a benchmark to aid improvement of this technology.

\noindent3) We improve the accessibility of this research by adapting the tasks to the reduced workspace and degrees of freedom of the DOBOT MG400, an inexpensive and commercially-available 4-DoF robot arm. Additionally, we commit to openly releasing all software and hardware developments in this work to aid dissemination of this benchmark.

\section{RELATED WORK} \label{sec:related}


\subsubsection{Deep Reinforcement Learning in Tactile Robotics}
Deep reinforcement learning (RL) has proved successful in solving many sequential decision making problems in robotics, particularly those with high-dimensional observation spaces such as in computer vision \cite{sunderhauf2018limits}. Some work has explored tactile RL with low-resolution tactile sensing to perform tasks such as object stabilisation \cite{van2016stable} and learning a forward predictive model \cite{veiga2017tactile}. Work in \cite{tian2019manipulation} proposed a deep tactile model-predictive RL framework for learning how to re-position an object using a Gelsight-style tactile sensor, and \cite{dong2021tactile} used Twin Delayed DDPG (TD3) \cite{fujimoto2018addressing} to learn a general tactile-guided insertion policy in the physical environment from a sequence of tactile images. A follow-up study simplified the observation space containing the raw tactile images and robot proprioceptive data for the deep RL policy by learning an extrinsic contact line model for contact localization. This policy was learned in a simulation environment \cite{kim2021active}. 

Recently, a tactile simulation environment for deep RL, called Tactile Gym \cite{church_tactile_2021}, has successfully applied the trained policies to some challenging tasks such as object pushing and rolling in a real physical environment.  The transfer from simulation to real-world physical environment was facilitated using a novel real-to-sim tactile image translation technique, in a zero-shot manner. Tactile-based deep RL has also been successfully applied to robotic service tasks like learning to type on a braille keyboard \cite{church2020deep} with the TacTip and learning to play the piano \cite{xu2021towards} with the DIGIT optical tactile sensor. 

\subsubsection{Tactile Sim-to-real Transfer}

The tactile sim-to-real gap significantly hinders the application of learned policy in simulation to reality. Two research directions have been explored to close this gap: using the Finite Element (FE) method to model the sensor deformation dynamics \cite{narang2021interpreting, narang2021sim, sferrazza2020learning, sferrazza2020sim, bi2021zero, Ding2020Sim-to-RealSensing}, or leveraging the image rendering method to replicate the sensory data \cite{gomes2021generation, wang2022tacto, church_tactile_2021, si2022taxim}. For a more thorough review we refer to \cite{church_tactile_2021}. In the present work, we follow the tactile sim-to-real method described in \cite{church_tactile_2021} by using depth image-rendering and image translation for two popular optical tactile sensor classes: GelSight-type which is based on image shading \cite{abad2020visuotactile} and TacTip type which is based on biomimetic marker-based transduction \cite{lepora2021soft}.

\section{Method and Experiments} \label{sec:method}

\subsection{Tactile Robot System}\label{subsec:TS}

In this paper, we use a tactile robot comprising a low-cost desktop robot arm with high-resolution tactile sensor mounted as an end effector. This is intended to be a lower-cost, desktop version of the setup used by Church et al~\cite{church_tactile_2021} for investigating sim-to-real tactile deep RL, which used a 6-axis industrial robot from Universal Robotics. In this paper, we expand the approach to compare three distinct optical tactile sensors: the TacTip, DIGIT, and DigiTac. Table \ref{table:plat_comp} compares this robot platform to those used in related works, in terms of the features, cost and the integration of tactile sensors. The robot platform and the operation of the tactile sensors are presented in detail in an accompanying second paper, with the main focus of the present paper on the application of sim-to-real deep RL on this low-cost desktop robot. Hence, we summarize just the main aspects of the tactile robot.

\subsubsection{Desktop robot arm} We use a Dobot~MG400 4-axis arm designed for affordable automation. The base and control unit has footprint 190\,mm$\times$190\,mm, payload 750\,g, maximum reach 440\,mm and repeatability $\pm0.05\,$mm. As we describe later, the accuracy of tactile models trained using this arm is similar to larger industrial robot arms. The main constraint is that only the $(x,y,z)$-position and rotation around the $z$-axis of the end effector are actuated.

\subsubsection{High-resolution optical tactile sensing} Here we consider three distinct optical tactile sensors: \\
\noindent (a) The {\bf TacTip}, a soft, curved, 3D-printed tactile skin with an internal array of pins tipped with markers, which are used to amplify the surface deformation from physical contact against a stimulus. For more details, we refer to ref.~\cite{ward2018tactip,lepora2021soft}.\\ 
(b) The {\bf DIGIT} shares the same principle of the Gelsight tactile sensor \cite{yuan2017gelsight}, but can be fabricated at low cost and is of a size suitable for integration of some robotic hands, such as on the fingertips of the Allegro hand~\cite{lambeta2020digit}.\\ 
(c) The {\bf  DigiTac} is an adapted version of the DIGIT and the TacTip, whereby the 3D-printed skin of a TacTip is customized to integrated onto the DIGIT housing, while keeping the camera and lighting system. In other words, this sensor outputs tactile images of the same dimension as the DIGIT, but with a soft biomimetic skin like other TacTip sensors.

\begin{table}
\vspace{-1em}
\addtolength{\tabcolsep}{-1pt}
\centering
\caption{Comparison of the existing tactile sim-to-real robotic platforms for TacTip-style (T) and GelSight-style (G) sensors, respectively.}
\begin{tabular}{@{}c|c|c|c|c|c@{}} 
\hline
\textbf{Platform}                                                                                                       & Features        & \cite{church_tactile_2021}        & \cite{Ding2020Sim-to-RealSensing}          & \cite{si2022taxim}       & Ours           \\ 
\hline
\multirow{7}{*}{\textbf{Robots}}                                                                                        & Type            & UR5        & Sawyer        & UR5        & MG400          \\
                                                                                                                                          & Accuracy~(mm)   & 0.1        & 0.1           & 0.1        & \textbf{0.05}  \\
                                                                                                                                          & Price~(\$)   & 45k       & 26k            & 45k       & \textbf{2.7k}   \\
                                                                                                                                          & Max.~Reach~(mm) & 850        & \textbf{1260} & 850        & 440            \\
                                                                                                                                          & Playload~(kg)   & \textbf{5} & 4             & \textbf{5} & 0.75           \\
                                                                                                                                          & Weight~(kg)     & 18.4       & 19            & 18.4       & \textbf{8}     \\
                                                                                                                                          & DoF             & 6          & \textbf{7}    & 6          & 4              \\ 
\hline
\multirow{3}{*}{\begin{tabular}[c]{@{}c@{}}\textbf{Sensors}\\\textbf{(if~integrated)~ }\end{tabular}} & TacTip~(T)      & \checkmark  & \checkmark     & \xmark      & \checkmark      \\
                                                                                                                                          & DIGIT~(G)       & \xmark      & \xmark         & \checkmark  & \checkmark      \\
                                                                                                                                          & DigiTac~(T)     & \xmark      & \xmark         & \xmark      & \checkmark      \\
\hline
\end{tabular}
\label{table:plat_comp}
\end{table}

\begin{figure*}[t]
  \centering
    \includegraphics[width=1\linewidth]{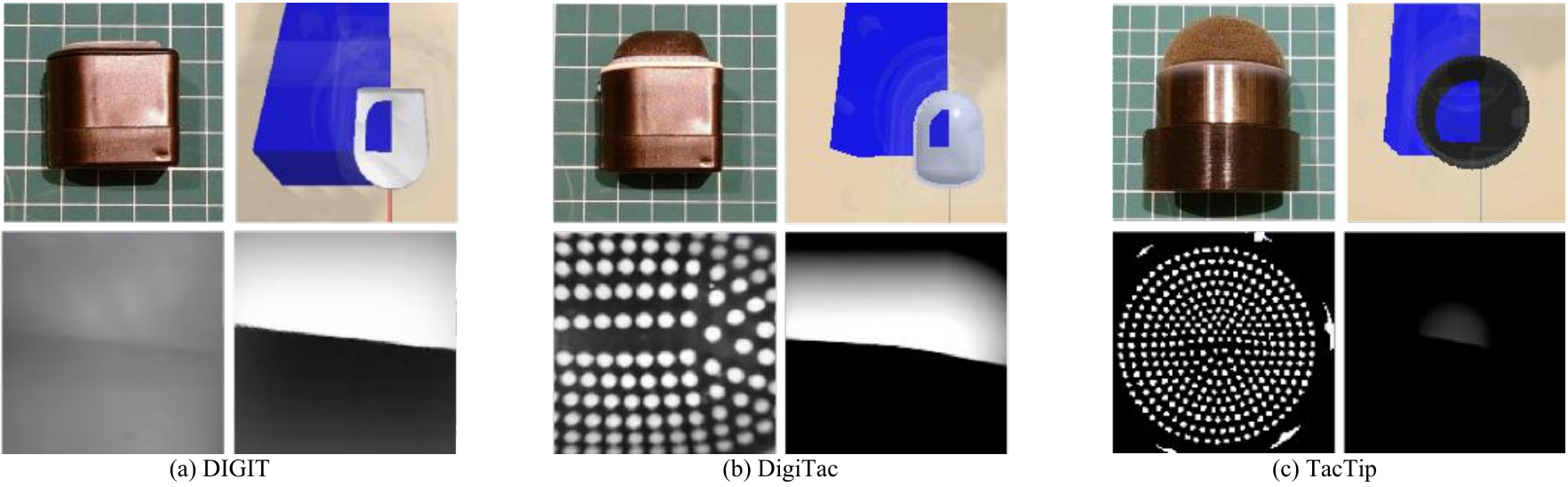}
    \caption{Comparison between the different optical tactile sensors: (a) DIGIT, (b) DigiTac, (c) TacTip. For each sensor we show: top-left, the real sensor hardware; top-right, simulated contact geometry between sensor skin and a blue edge stimulus; bottom-left, tactile image gathered by the real sensor pressing onto an edge; and bottom-right, a generated depth image matching those gathered in simulation.}
  \label{fig:real_sim_t_img}
\end{figure*}


\subsection{Tactile Sim-to-Real Deep RL Framework}\label{subsec:TS2R}

Following Church et al~\cite{church_tactile_2021} we take a sim-to-real deep reinforcement learning approach to achieve the desired robot behaviour on physical hardware. This approach consists of three distinct phases:\\ \textbf{1)} An online learning in simulation phase, where deep RL is applied to images captured by a simulated tactile sensor for the learning of several distinct tasks (here edge following, surface following and object pushing).\\ \textbf{2)} A domain adaption phase where a model is learned for the translation of images captured by a real tactile sensor to images captured by the simulated sensor.\\ \textbf{3)} A zero-shot sim-to-real phase where policies learned in simulation are transferred to the real hardware using the networks trained in the previous two steps. An overview of this approach is shown in Fig.~\ref{fig:overview} and more details can be found in the original reference~\cite{church_tactile_2021}.

To adapt the existing approach\cite{church_tactile_2021} to our setting, we needed to make several changes. In this work we chose to use a low-cost desktop robot arm as described above (Sec.\ref{sec:method} A). This arm has only 4 actuated axes, whereas previous work used a 6-axis industrial robotic arm (UR5, Universal Robotics). To facilitate the use of this lower degree-of-freedom arm we have adapted the tasks and data collection procedures while attempting to meet the challenge of successful performance on tasks originally developed for a more capable robot arm. In particular the previous surface following experiments made use of Roll and Pitch to accurately maintain a normal orientation to a surface. Instead, we rely on a custom 3D-printed flange so the sensor is mounted perpendicular to the end effector. In this way, we can make use of the Yaw DoF of the DOBOT MG400 when maintaining normal orientation to a surface varying in only one direction, as shown in Fig.~\ref{fig:overview}(d).

Moreover, whilst it was previously claimed that the approach used in \cite{church_tactile_2021} should be applicable to ``a broad range of other high-resolution optical tactile sensors, including sensors of the Gelsight type'', it was only demonstrated with a hemispherical and flat TacTip. Here we validate that the approach works with the DIGIT sensor of the Gelsight-type, which has a distinct tactile sensing principle based on image shading in place of marker-based transduction. We also validate the approach with alternative forms of the TacTip sensor, including a new (more compact) version of TacTip than used in \cite{church_tactile_2021} and the DigiTac which has a TacTip skin on a DIGIT housing. When compared with the original work these sensors introduce differences in camera perspectives, lens dynamics, marker size, density of markers and have significantly different lighting conditions. To do this, we extended the simulation to include simulated sensors that matched real hardware using the CAD files for 3D-printing those sensors.

\subsection{Deep RL with Different Tactile Dynamics in Simulation}\label{subsec:method_task}
The sim-to-real deep RL framework uses the Tactile Gym~\cite{church_tactile_2021} (see above) to simulate the contact dynamics with rigid-body physics, using tactile information rendered as depth images relative to the CAD model for the sensor. Thus, the learnt tactile-feedback policies from~\cite{church_tactile_2021} apply only the TacTip originally considered in that study. 

Hence, we extend the Tactile Gym \cite{church_tactile_2021} with two new virtual optical tactile sensors: DIGIT (Gelsight-style) and DigiTac (TacTip-style) (Fig.~\ref{fig:real_sim_t_img} a,b), based on their open-source CAD files \cite{lambeta2020digit} used for 3D-printing the DIGIT. We follow the method described in \cite{church_tactile_2021} to efficiently capture the depth images as tactile images by synthetic cameras embedded within those sensors. Specifically, we adjust the physics (friction, stiffness, damping, etc.) of the skin and the parameters of the camera for each sensor, to achieve realistic performance during learning.

The three distinct tactile observation spaces corresponding to the three type of sensor (Fig. \ref{fig:real_sim_t_img}) are then used as input to train deep RL polices for the tasks, following the methods detailed in~\cite{church_tactile_2021}. We report on the training results in Sec. \ref{sec:exps_ressults}.

\subsection{Sim-to-real Transfer for Tactile Images}\label{subsec:sim2real}

Here we aim for zero-shot learning so the learned policy in simulation can be transferred to the real-world task without further training or tuning. Hence it is essential to have a model to bridge the gap between the simulated and real domains. Progress in Generative Adversarial Network (GAN) methods has enabled realistic image generation, which we leverage to learn an image-to-image translation GAN \cite{isola2017image} applied to real-to-sim tactile image translation. The network takes advantage of a U-net architecture \cite{ronneberger2015u} for the image-conditioned generator to infer better the spatial features during training, alongside a standard convolutional network \cite{krizhevsky2012imagenet} with batch normalization \cite{ioffe2015batch} for the discriminator. 

Since each of the three optical tactile sensors considered here have a different design and illumination, the image-preprocessing and hyper-parameters need tuning for each sensor. The arrays of markers in the TacTip and the DigiTac reflect light more clearly, which eases the fine-tuning of their image processing compared to the DIGIT, where the image shading is more subtle. However, after tuning we find that all sensors work effectively for sim-to-real transfer, which we will cover in the results sections later.


\begin{figure*}[ht]
  \centering
     \includegraphics[width=0.95\linewidth]{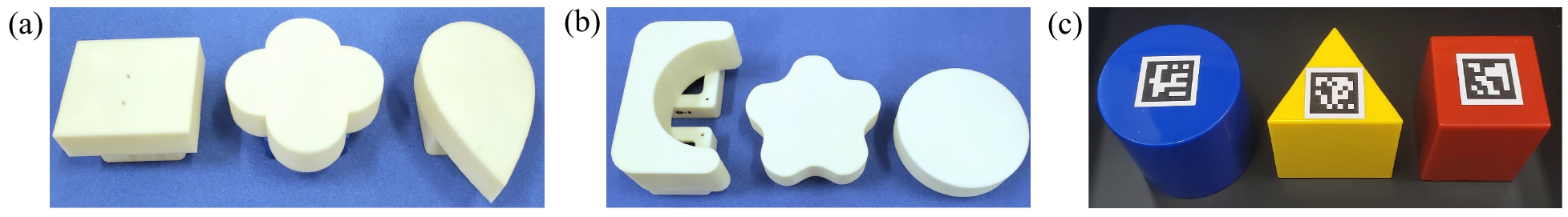}
    \vspace{-1em}
    \caption{Tactile stimuli used in manipulation tasks: (a) square, clover leaf, and teardrop stimuli for the edge following task; (b) arch, flower, and disc for the (side) surface following task; (c) blue cylinder, yellow triangular prism, and red cube for the object pushing task (ArUco markers for ground truth tracking).}
  \label{fig:objects}
\end{figure*}

\begin{figure*}[b!]
  \centering
\vspace{-1em}
     \includegraphics[width=1\linewidth]{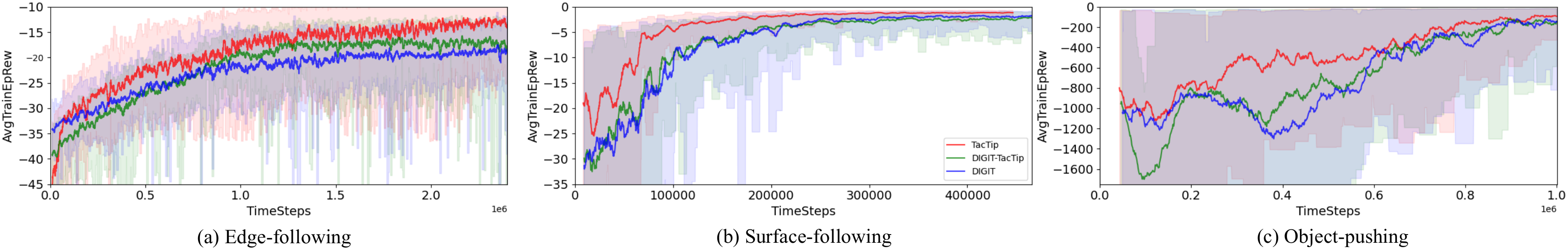}
     \vspace{-2em}
    \caption{Averaged training performance of learned policies for different sensors in (a) edge-following, (b) surface-following, (c) object-pushing tasks.}
  \label{fig:rl_results}
\end{figure*}

\begin{table}[h]
\addtolength{\tabcolsep}{-1pt}
\caption{Sensor pose sampling ranges used during data collection. Sensor poses are expressed relative to a fixed coordinate frame attached to the training feature ($Rz$ = axial rotation around $z$-axis).}
\centering
\begin{tabular}{c|c@{}c@{}c|c@{}c@{}} 
\hline
\textbf{Contact~Features}                 & \multicolumn{3}{c|}{\textbf{Edge }} & \multicolumn{2}{c}{\textbf{Surface }}  \\ 
\hline
\diagbox{\textbf{Sensors}}{\textbf{Axis}} & $y$\,(mm)\, & \,$z$\,(mm)\,\,\,   & $Rz$\,(deg)       & $x$\,(mm)\,\,\,   & $Rz$\,(deg.)                     \\ 
\hline
\textbf{Tactip}                           & {[}-6,6] & {[}2,5] & {[}-179,180]~ & {[}1,4] & {[}-15,15]~                 \\
\textbf{DigiTac}                          & {[}-5,5] & {[}2,4] & {[}-179,180]~ & {[}1,3] & {[}-15,15]~                 \\
\textbf{DIGIT}                            & {[}-5,5] & {[}2,3] & {[}-179,180]~ & {[}1,2] & {[}-11,11]~                 \\
\hline
\end{tabular}
\label{table:ranges}
\end{table}

\subsection{Sim-to-real Data Collection}\label{subsec:RL}
The three manipulation tasks considered here require distinct sim-to-real models across two distinguishing contact features: an edge for the edge-following task, and a surface for the surface-following and object-pushing tasks. We collect a training and validation dataset per contact feature (edge and surface) and per sensor, leading to twelve datasets in total.


Each training dataset comprised 5000 tactile images and each validation dataset 2000 tactile images, collected by using the desktop tactile robot (Sec.\ref{sec:method} A) to randomly contact data of the appropriate edge or surface feature. For the data collection, straight edge and flat surface of a 3D-printed stimulus, labeled with the  relative poses between the sensor and the stimulus under contact. The movement ranges for each random contact feature are shown in Table \ref{table:ranges}. These datasets take about 6 hours to collect on the physical robot and less than 1 minute to collect in simulation.

A further subtlety for the DIGIT and the DigiTac is that they are no longer symmetric (being broader across one axis) unlike the original TacTip. This meant we needed to customize the range of $(x,y)$-pose data collection depending on the rotation angle, which was implemented by scaling the $y$-range orthogonal to the edge by the tangent of the angle. 


\subsection{Tactile Control Tasks}\label{subsec:method_task}

Here we adapt three tactile control tasks proposed in \cite{church_tactile_2021} to the desktop tactile robot: edge-following, surface-following and object-pushing. Although we expect our platform would also be viable for the ball-rolling task, we do not implement it in this work because the DigiTac does not currently have a flat skin that is suitable for ball-rolling. 
\subsubsection{Object Pushing} This task aims for the robot to push an object through a sequence of goal positions along a trajectory on a flat surface. Three trajectories are considered: straight, curved and sinusoidal. In practise, each trajectory was divided into 10 equal-length sections with the final point on each section specified as the goal position; thus, there are ten goals for each trajectory. The 2D action space comprises the $x$-position and rotation angle of the tool center point (TCP) located at the tip of the tactile sensor. We use three distinct objects with different shapes and weights (Fig. \ref{fig:objects}c): a triangular prism, cube and cylinder varying in weight from 185\,g to 363\,g. These objects differ from those used in \cite{church_tactile_2021}, which were lighter of mass (50\,g to 80\,g), to better suit the elasticity of the DIGIT that required a heavier object for an appreciable tactile deformation. An ArUco marker is place on top of each object to track its movement for comparison with the ground truth using the tracking method described in \cite{lloyd2021goal}.



\subsubsection{Edge Following} This task aims for the robot to slide the tactile sensor along a contacted edge while maintaining a fixed distance between the edge and the TCP located at the tip of the tactile sensor. The 2D action space comprises the $x$ and $y$ position of the TCP. To evaluate the robustness and maneuverability of the tactile robot, we use three stimuli that contain interesting geometrical features such as straight edges, positively/negatively curved edges and a right-angled corner (Fig. \ref{fig:objects}a). To measure tracking performance, the ground truth shapes are extracted from the CAD models of these 3D-printed objects by importing into the Blender CAD software \cite{blender} and outputting the point clouds of their boundaries.

\subsubsection{Surface Following} This task aims for the robot to slide the tactile sensor over a contacted surface while maintaining a fixed contact depth and orientating the TCP representing the tip of the sensor normal to that surface. The 2D action space comprises the $y$-position and rotation angle of the TCP. To evaluate the robustness and maneuverability of the tactile robot, we use the side surfaces of three objects comprising closed-loop 2D surfaces in 3D space (Fig.~\ref{fig:objects}b) that contain interesting geometrical features such as locally planar, concave and convex surfaces. As with the edge-following task, the ground truth shapes and dimensions of the 3D-printed stimuli are obtained from the CAD models using Blender.

%
\section{Results}\label{sec:exps_ressults}
 
\begin{figure}[t!]
  \centering
     \includegraphics[width=1\linewidth]{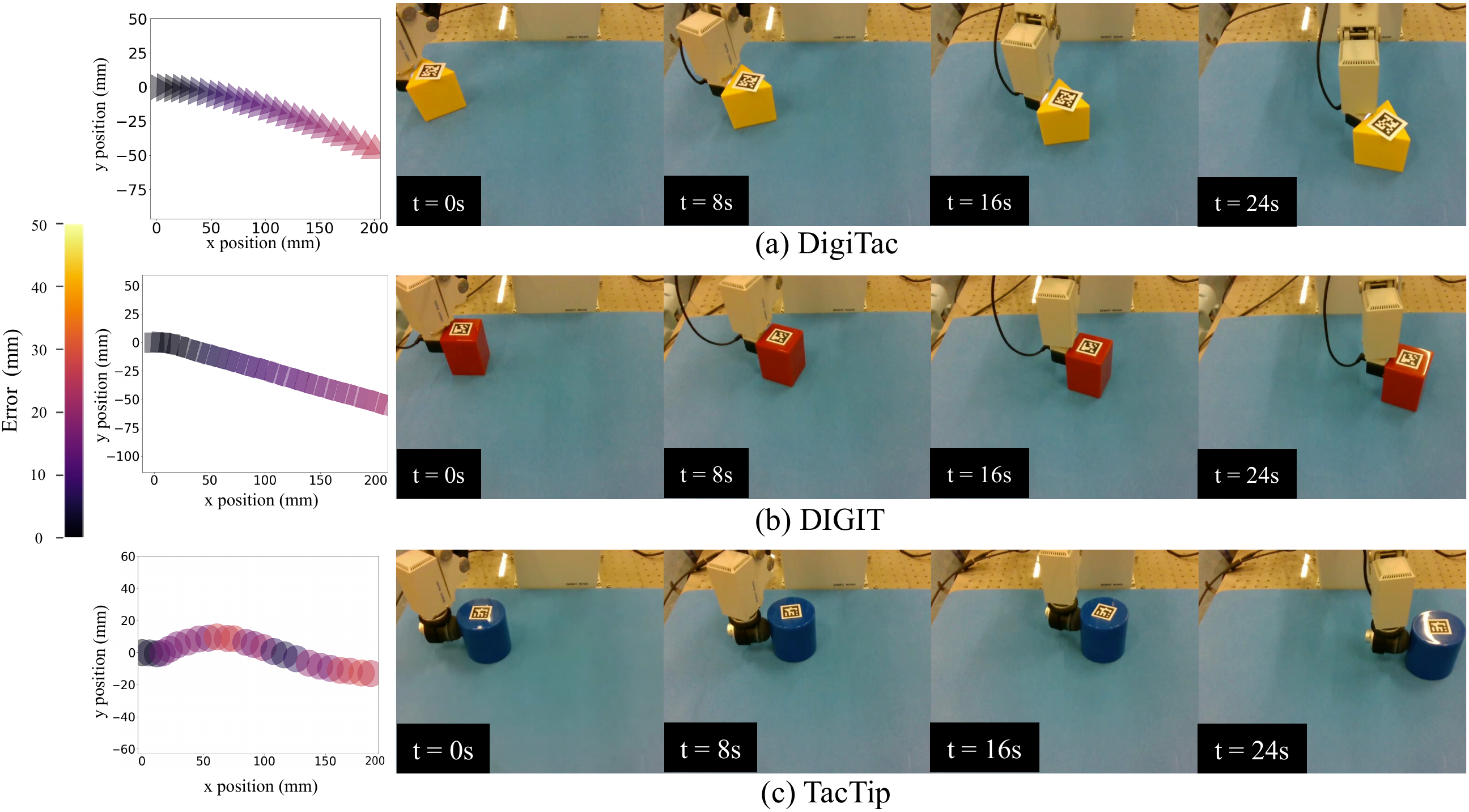}
     \vspace{-2em}
    \caption{The tactile robot executing 3 pushing policies: (a) triangular prism along a curve trajectory with the DigiTac, (b) cube along a  straight line with the DIGIT, (c) cylinder along a sinusoidal trajectory with the TacTip. The plots in the most left columns are the objects actual paths recorded by the tracking system; the right four columns are snapshots taken from the camera.}
  \label{fig:push_exp}
\end{figure}

\subsection{RL Performance in the Simulated Environments}

The deep RL method PPO~\cite{schulman2017proximal} is here used to learn policies for the three simulated optical tactile sensors on the three considered tasks (pushing, edge-following, surface-following). Specifically, we use the Stable-Baselines-3 \cite{raffin2019stable} implementation of PPO for training the policies. 

The results of training in simulation are given in Fig.~\ref{fig:rl_results}. Although there are slight differences in performance during training between the sensors, all the learned policies achieved similar accurate final performances in all tasks. We attribute these small differences to the different contact dynamics due to the different shapes of the tips for the three tactile sensors, but as we mentioned this does not affect the overall performance. In particular, we did not find it necessary to fine-tune the training hyperparameters for each tactile sensor. In addition, we apply the learned policy in each simulated task to the real robot later without any further fine-tuning.

\begin{table}[h!]
\vspace{-1em}
\addtolength{\tabcolsep}{-1pt}
\caption{Mean SSIM values for image translation GANs trained with edge and surface features (values closer to 1.0 represent better image translation).}
\centering
\begin{tabular}{l|ccc} 
\hline
\diagbox{\textbf{Features}}{\textbf{Sensors}} & \textbf{Tactip} & \textbf{DigiTac} & \textbf{DIGIT}  \\ 
\hline
\textbf{Edge}                                 & 0.9956          & 0.9953           & 0.9867          \\ 
\textbf{Surface}                              & 0.9927          & 0.9932           & 0.9818          \\ 
\hline
\end{tabular}
\label{table:ssim}
\end{table}
\subsection{Image Translation for Sim-to-real Transfer}

We use the Structural Similarity Index (SSIM) to evaluate the quality of the translated images for both the edge and surface validation datasets (Table~\ref{table:ssim}). The high SSIM values close to unity (perfect match) obtained with the trained model shows that the image-to-image translation produces accurate tactile images for all sensors. Therefore, the method implemented from \cite{church_tactile_2021} is applicable not only for TacTip family of tactile sensors but also for GelSight-style sensors such as the DIGIT. Some examples of tactile images and their translated versions for edge contacts are shown in Fig. \ref{fig:real_sim_t_img}. 

As an aside, we comment that the performance of the DIGIT image translation was slightly lower than the TacTip and DigiTac. It is possible that a different architecture of neural network more suited to the GelSight-type sensors could change this result. We note that this minor difference does not affect the task overall performance, which we attribute to the deep RL policies being learnt with domain randomization.

\begin{table*}[t!]
\addtolength{\tabcolsep}{-1pt}
\caption{Mean Euclidean distances of the actual trajectories from the ground-truth trajectories for the object pushing task. The numbers in bold denote the best result among the three sensors for the same object and trajectory. Failure cases are denoted "N/A". All the experiments are using the objects without additional weights except the final row "DIGIT $\backslash$ (weighted)" where we increase all the objects' weights by 150\,g.}
\centering
\begin{tabular}{c|ccc|ccc|ccc} 
\hline
\textbf{Trajectories}                      & \multicolumn{3}{c|}{\textbf{Straight }}                                                 & \multicolumn{3}{c|}{\textbf{Curve }}                                                                                     & \multicolumn{3}{c}{\textbf{Sine }}                                                                \\ 
\hline
\diagbox{\textbf{Sensors}}{\textbf{~Obj.}} & \textbf{Cube}               & \textbf{Cylinder}           & \textbf{Tri.~Prism}         & \textbf{Cube}               & \textbf{Cylinder}           & \textbf{Tri.~Prism}                                          & \textbf{Cube}                        & \textbf{Cylinder}           & \textbf{Tri.~Prism}          \\ 
\hline
\textbf{Tactip}                            & \textbf{10.33~mm}          & \textbf{9.21~mm}           & \textbf{11.51~mm}          & \textbf{12.19~mm}          & 11.29~mm                   & \begin{tabular}[c]{@{}c@{}}\textbf{13.20 mm}\\\end{tabular} & \textbf{11.93~mm}                            & \textbf{11.30~mm}          & \textbf{13.89~mm}           \\ 
\hline
\textbf{DigiTac}                           & 11.25~mm                   & 10.24~mm                   & 16.09~mm                   & 13.01~mm                   & \textbf{11.20 mm}          & 16.41~mm                                                    & 12.32~mm                            & 11.48~mm                   & 15.13~mm                    \\ 
\hline
\textbf{DIGIT}                     & 11.20~mm                   & 10.13~mm                   & {N/A}                        & 12.94~mm                   & 12.00~mm                   & {N/A}                                                         & 12.41~mm                            & 11.33~mm                   & {N/A}                         \\ 
\hline
\textbf{DIGIT $\backslash$ (weighted)}   & 10.92~mm & 11.00~mm & 16.65~mm & 12.28~mm & 11.51~mm & 16.58~mm                                  & 12.07~mm & 11.53~mm & 17.06~mm                         \\ 
\hline
\end{tabular}
\label{table:push_eval}
\end{table*}
\begin{table}[t!]
\vspace{-1em}
\addtolength{\tabcolsep}{-1pt}
\caption{Mean Euclidean distances of the trajectories from the ground truth for the specific object (triangle prism) pushing task using DIGIT. The number in bold denotes the best result among the weights.}
\centering
\begin{tabular}{c|c|c|c} 
\hline
\diagbox{\textbf{Weights}}{\textbf{~Traj.}} & \textbf{Straight}  & \textbf{Curve}     & \textbf{Sine}        \\ 
\hline
\textbf{185g + 0g}                                            & N/A                & N/A                & N/A                  \\
\textbf{185g + 50g}                                           & 18.05~mm          & 17.73~mm          & 17.95~mm            \\
\textbf{185g + 100g}                                          & 16.93~mm          & 17.54~mm          & 18.12~mm            \\
\textbf{185g + 150g}                                          & \textbf{16.65~mm} & \textbf{16.58~mm}          & \textbf{17.06~mm}            \\
\textbf{185g + 200g}                                          & 17.17~mm          & 16.92~mm & 17.26~mm        \\ 
\hline
\end{tabular}
\label{table:DIGIT_tri_push_exp}
\end{table}

\subsection{RL Performance in the Physical Environments}

\subsubsection{Object Pushing}

We consider first the object pushing task, where the tactile robot must move the object along a desired trajectory through a series of goal points. Successful task performance corresponds to accurately pushing the test object (a cube, cylinder or triangular prism) along the trajectories (straight: $y = kx$, curved: $y = 0.001x^2$, sinuosoidal: $y = 0.02\sin(0.02x)$, where $k\in[-0.3,0.3] ,x\in [0,200]$\,mm), using feedback from the tactile sensor to maintain the object on its desired path. 

The tactile robot successfully pushed the object along its desired trajectory, with a typical mean Euclidean distance the actual trajectory from its intended trajectory of $\sim$10\,mm (Table \ref{table:push_eval}), compared to an overall distance travelled of 200-280\,mm (250 steps in total for each episode) and a sensor tip size of 20-40\,mm. The successful task performance is also indicated by snapshots taken from the trajectories (Fig.~\ref{fig:push_exp}) and videos are provided in supplementary material. Overall, the performance when successfully pushing the objects is similar to that reported in \cite{church_tactile_2021}. 

Examining the results more closely, the TacTip performs slightly better than the DigiTac with accuracies of 9-13\,mm compared with 11-16\,mm, which we attribute to a stabler push due to the larger contact surface. The DIGIT has similar accuracy to the DigiTac, but failed at the pushing task with the triangular prism for all trajectories. We hypothesise that the main reason of this failure is that the triangular prism is the lightest object (185g) and the DIGIT has a relatively stiff elastomer compared with the TacTip and DigiTac, which causes the tactile image translation to fail on this object. To validate this hypothesis, we extend the experiments with the DIGIT to pushing the triangle prism with additional weights (ranging from 50-200\,g) using the same deep RL and GAN models. The results (Table \ref{table:DIGIT_tri_push_exp}) show that the performance improves with increased object weight upto 150\,g after which there was no benefit. We expect this is because the additional weight helps the tactile images lie within the distribution of the GAN training for accurate real-to-sim transfer. Once in this range, the bottleneck on the performance is the RL policy instead of the GAN.

\subsubsection{Edge Following}
Next, we consider the edge following task, where the tactile robot must slide the sensor around the edge of various planar objects with geometrical features such as curved edges and a corner. We note that the sim-to-real image translation was trained only on a straight edge, but as we see below the method generalizes to more complex shapes. 

In all cases, the tactile robot successfully completed the edge-following trajectory, with typical mean Euclidean position errors of 0.6-1.4\,mm for the DigiTac and TacTip, and 0.9-1.8\,mm for the DIGIT (Table~\ref{table:edge_eval}). The successful task performance is shown by the trajectories superimposed on the ground truth shapes (Fig.~\ref{fig:edge_exp}) and videos are provided in supplementary material. Again these results are comparable to those reported in \cite{church_tactile_2021} and also for servo control under supervised learning of the pose~\cite{lepora2021pose}.

Examining the results more closely, the DIGIT can achieve accurate performance (mean Euclidean position errors of 1.5\,mm overall) when traversing the edge contours despite having some regions of larger error (Fig.~\ref{fig:edge_exp} top row, coloured regions up to 5\,mm error). This is because the flatter stiffer elastomer of the DIGIT sensing surface causes an increased sensitivity to small deviations in penetration distance while moving around the object, relative to the softer TacTip and DigiTac sensing surfaces. For sufficient deformation of the DIGIT sensor to give good performance, a more forceful contact needs to be applied, which also increases the frictional force. To avoid damaging the sensor, we mitigate this friction by coating the objects with wax to ease the sliding motion.


\begin{figure}[t!]
  \centering
  \vspace{0em}
     \includegraphics[width=1\linewidth]{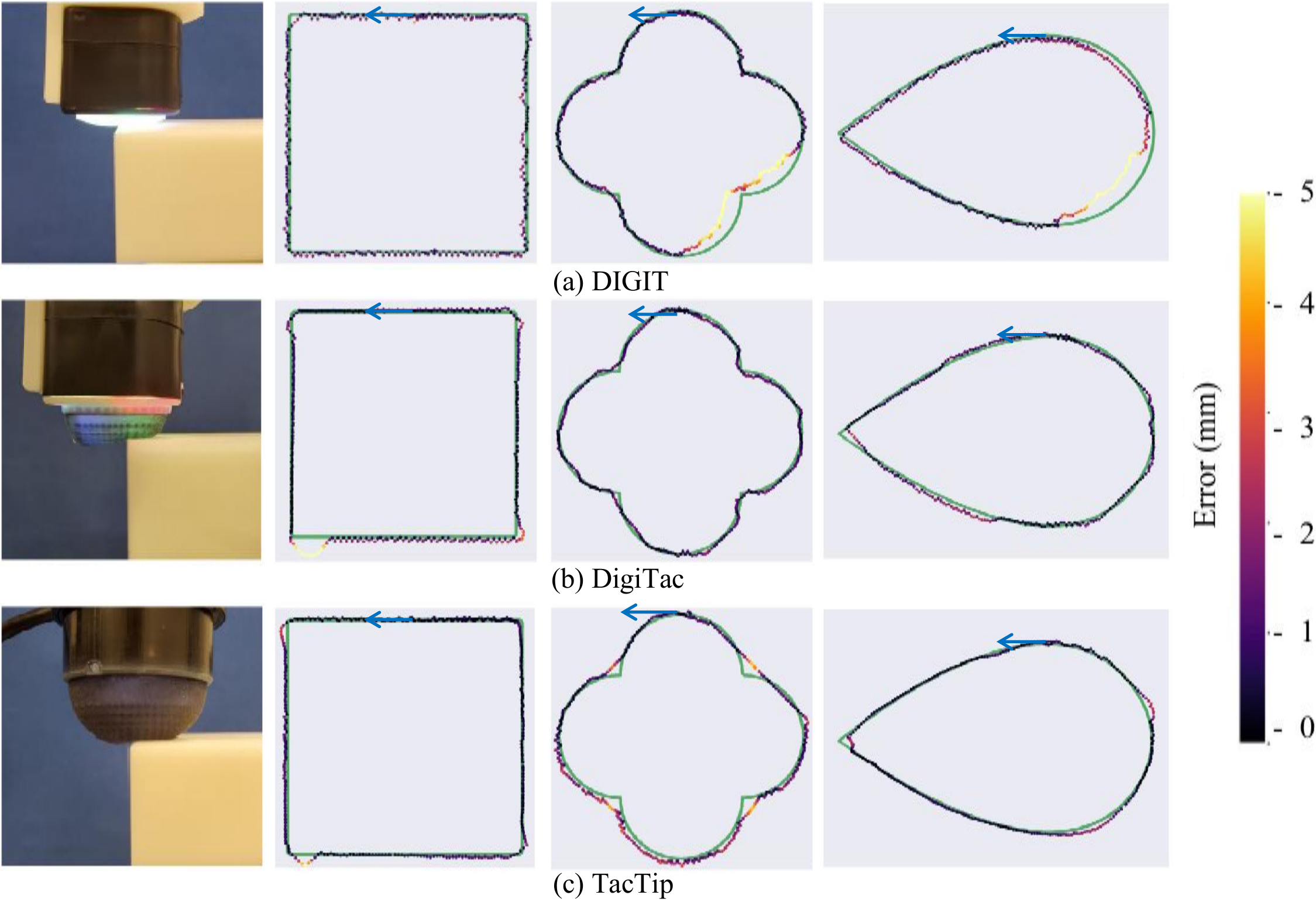}
     \vspace{-1.5em}
    \caption{The tactile robot executing edge-following policies on 3 distinct shapes for the (a) DIGIT, (b) DigiTac and (c) TacTip. The ground truth for each object is shown in green and errors of the traced contour from the ground truth are colour-coded (side colour bar). The blue arrow denotes the starting point and direction.}
  \label{fig:edge_exp}
    \centering
  \vspace{2em}
  \includegraphics[width=1\linewidth]{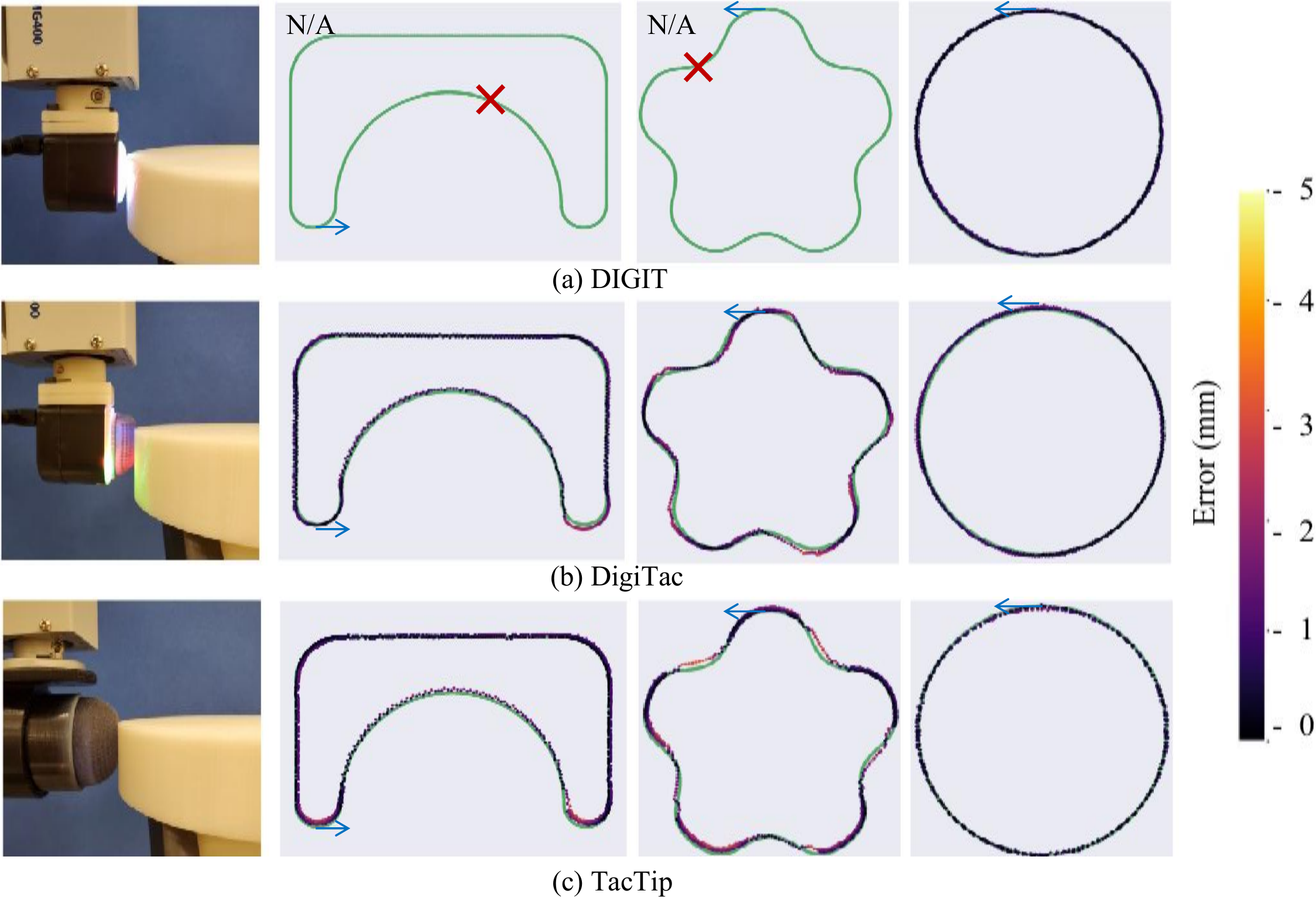}
     \vspace{-1.5em}
    \caption{The tactile robot executing 3 surface-following policies on 3 distinct shapes, corresponding to the (a) DIGIT, (b) DigiTac and (c) TacTip. The ground truth for each object is shown in green and errors of the traced contour from the ground truth are colour-coded (side colour bar). The blue arrow denotes the starting point and direction. The DIGIT failed to follow the arch and flower objects at points denoted by the red crosses.}
  \label{fig:surf_exp}
\end{figure}

\begin{table}[t!]
\vspace{0em}
\addtolength{\tabcolsep}{-1pt}
\caption{Mean Euclidean distances of the trajectories from the ground truth for the edge following task. The best result from the three sensors is shown in bold.}
\centering
\begin{tabular}{l|l|l|l} 
\hline
\diagbox{\textbf{Sensors}}{\textbf{Obj.}} & \textbf{Square}            & \textbf{Clover}   & \textbf{Foil}      \\ 
\hline
\textbf{DIGIT}                            & 0.88~mm                   & 1.71~~mm         & 1.82~mm           \\
\textbf{DigiTac}                     & 1.04~mm                   & \textbf{0.85~mm} & 0.86~mm           \\
\textbf{TacTip}                           & \textbf{0.63}~\textbf{mm} & 1.42~mm          & \textbf{0.67~mm}      \\ 
\hline
\end{tabular}
\label{table:edge_eval}
\vspace{2em}
\addtolength{\tabcolsep}{-1pt}
\caption{Mean Euclidean distances of the actual trajectories from the ground truth trajectories for the surface following task. The number in bold denotes the best result from the three sensors. Failure cases are indicated by "N/A".}
\centering
\begin{tabular}{l|l|l|l} 
\hline
\diagbox{\textbf{Sensors}}{\textbf{Obj.}} & \textbf{Arch}     & \textbf{Flower}   & \textbf{Circle}    \\ 
\hline
\textbf{DIGIT}                            & N/A               & N/A               & \textbf{0.47~mm}  \\
\textbf{DigiTac}                     & \textbf{0.79~mm} & \textbf{1.04~mm} & 0.58~mm           \\
\textbf{TacTip}                           & 0.91~mm          & 1.23~~mm         & 0.59~mm    \\ 
\hline
\end{tabular}
\label{table:surf_eval}
\end{table}


\subsubsection{Surface Following}

Finally, we consider the surface following task, where the tactile sensor must slide around the curved surfaces of various objects with vertical walls, which have geometrical features such as concave and convex surfaces. We note that the sim-to-real image translation was trained only on a planar surface, but as we see below the method generalizes to more complex shapes. 

The tactile robot successfully completed the surface-following trajectory, with typical mean Euclidean position errors of 0.6-1.2\,mm for the DigiTac and TacTip, and the DIGIT giving the most accurate trajectory of 0.5\,mm error on the circular wall (compared to 0.6\,mm for the other sensors). The successful task performance is shown by the trajectories superimposed on the ground truth shapes (Fig.~\ref{fig:surf_exp}) and videos are provided in supplementary material. Again these results are comparable to those reported in \cite{church_tactile_2021} and also for servo control under supervised learning of the pose~\cite{lepora2021pose}. 

During the experiments, we observed that the DIGIT tends to get stuck in the concave-shape surface (shown in the supplementary video). This is again because of the DIGIT's flatter, stiffer sensing surface which hinders its sliding movement over concave surfaces. To avoid breaking the sensor, we decided not to conduct this task with DIGIT on the flower and the arch. 
\section{DISCUSSION AND FUTURE WORK} \label{sec:discussion}

In this paper, we developed a low-cost tactile robot platform for sim-to-real deep reinforcement learning based on Tactile Gym \cite{church_tactile_2021}. The hardware included a desktop robot (DOBOT MG400) and three low-cost high-resolution optical tactile sensors: the TacTip, DIGIT, and DigiTac. We also integrated CAD models of the DOBOT MG400 and the considered sensors into the Tactile Gym and successfully learned policies for all sensors in several physically-interactive tasks involving different contact dynamics. To train an effective GAN model for real-to-sim image translation, we fine-tuned the image preprocessing parameters and calibrated the internal camera of each simulated tactile sensor so that the distribution of the simulated dataset was well-matched with the real dataset. 


The performance of our low-cost tactile sim-to-real deep RL robot platform was evaluated in three real-world physically-interactive tasks: edge-following, surface-following and object-pushing. The experimental results show that the developed platform is effective for all tasks with zero-shot performance on real objects or trajectories unseen in the simulation learning for all three tactile sensors. The main differences in performance between the sensors were due to the physical construction and material properties, rather than the different sensing mechanisms. For example, the flatter, stiffer construction the DIGIT with a GelSight-type sensing surface made it unsuited for following concave surfaces, unlike the soft domed structure of the TacTip and DigiTac.


Such empirical studies should help other researchers select and customize the appropriate physical characteristics of tactile sensors for different manipulation scenarios. Overall, we view the generality of our low-cost platform as opening up the possibility to apply either TacTip-style or GelSight-style tactile sensors to learning general sim-to-real deep RL policies for desired complex behaviors. The tactile robot platform should also benefit sim-to-real prehensile and dexterous manipulation tasks, for example by enabling the fundamental methods to be developed in controlled scenarios before applying them to more challenging applications with dexterous robot hands.

{\em Acknowledgements:} We thank Mike Lambeta and Roberto Calandra for donating the DIGIT sensors. We thank Di Wu for her preliminary work on the DOBOT hardware development. We thank Andrew Stinchcombe for
helping with the 3D-printing of the stimuli.

\bibliographystyle{unsrt}
\bibliography{manual}

\end{document}